\documentclass{article}

\usepackage{arxiv}

\usepackage[utf8]{inputenc} 
\usepackage[T1]{fontenc}    
\usepackage{hyperref}       
\usepackage{url}            
\usepackage{booktabs}       
\usepackage{amsfonts}       
\usepackage{nicefrac}       
\usepackage{microtype}      
\usepackage{lipsum}		
\usepackage{graphicx}
\usepackage{natbib}
\usepackage{doi}

\title{Semantic-based Self-Critical Training For Question Generation}

\author{ Loïc Kwate Dassi \\
	Ensimag, Grenoble, France \\
	\texttt{loic.kwate-dassi@grenoble-inp.org} \\
}

\hypersetup{
pdftitle={Semantic-based Self-critical Training For Question Generation},
pdfsubject={question generation},
pdfauthor={L. Kwate Dassi},
}

\begin{document}
\maketitle
\begin{abstract}
Question generation is a conditioned language generation task that consists in generating a context-aware question given a context and the targeted answer. Train language modelling with a mere likelihood maximization has been widely used while suffering from exposure bias and the discordance between the training and the test metrics. In the way of addressing this issue, The presented work portrays a fully Transformer-based reinforcement learning generator-evaluation architecture for neural question generation. To edge the flexibility of the generation, a semantic-based reward score was externally infused during the training to drive the training of the language model. The global architecture is laid out in a generator-evaluator fashion optimized directly to n-gram and semantic-based metrics. Evaluation metrics for language modelling only based on n-gram overlapping do not consider semantic relations between reference and candidate sequences. To improve the evaluation step, a two-fold evaluation was carried out. On the one side, an n-gram overlapping evaluation using the BLEU score. On the other side, a semantic-based assessment using BERTScore and NUBIA. The results were corroborated by a binary human evaluation of the semantic relatedness of the generated question and the ground truth. The results obtained showed that use a semantic-based REINFORCE algorithm for the question generation syntactically reshapes the generated questions while preserving their underlying semantic meaning. Many downstream applications can be drawn from a successful question generation including the enlargement of question answering datasets, the improvement of conversational systems, the enhancement of autonomous educational assessment systems, and so forth.
\end{abstract}

\section{Introduction}
Transformer architecture \citep{vaswani2017attention} has broken new ground in Natural Language Processing. Nowadays, it is widely used in a broad range of tasks including language understanding, language generation \citep{devlin2019bert, raffel2020exploring} as well as computer vision \citep{carion2020endtoend}. The question generation falls into the category of sequence-to-sequence task, thanks to the ability of Transformer architecture with positional encoding to approximate any continuous sequence-to-sequence transformations within compact support \citep{yun2020transformers}. Question generation is a sensitive task because it requires first a good understanding of the context, then requires how to establish a relation between the context and answer, and finally requires the model the ability to generate the questions fluently as a human. Through this work, an investigation of the potential of Transformer-based models on a question generation task is undertaken, and similar results close to the state-of-the-art have been achieved while keeping the model smaller. More precisely, the architecture holds 220M trainable parameters as opposed to the large architecture of Prophenet \citep{qi2020prophetnet} leading the state-of-the-art with 330M of trainable parameters. The neural networks developed in natural language processing, broadly in deep learning, become larger, these large models are energy demanding and are not widely accessible to the broad community. Investigating the aptness of the relatively small architectures to perform at the same stage of the larger architecture is noteworthy in the way of keeping the computational cost relatively small while pushing the boundaries of the state-of-the-art. This work is an extension of the work on Reinforcement Learning for question generation \citep{chen2020reinforcement} where the generator was trained through the REINFORCE algorithm. In the generator-evaluator architecture implemented, the generator is a sequence-to-sequence model Google’s T5 \citep{raffel2020exploring} and the evaluator is the mixture of BLEU score \citep{kishore2002bleu} and the cosine similarity of the classification token’s learning representation output by the Transformer-based model ELECTRA-discriminator \citep{clark2020electra}. The main long-term motivation of this work is the willingness to improve the education system by providing an autonomous system able to generate questions as humans, which can then be used to alleviate the conception of students’ examinations and support self-taught students in their studies.

\section{Related work}
\par 
Rule-based methods for question generation follow the directives set up by a specific expert. This range of models entirely relies on the knowledge of the expert who designs the rules of generating questions and the performance is limited by the expert's knowledge. Automatic Factual Question Generation from Text was introduced by \citep{heilman2020rulebasedquestiongeneration}. Specifically focused on generating factual WH-questions, the designed system produces questions by successively applying a set of manually encoded transformation rules to the statements.
\par 
\citep{zhou2017neural} introduced an RNN-GRU based encoder-decoder architecture that generates answer-aware questions. The encoder takes as inputs the context, the answer position, and linguistics features including part-of-speech (POS) and named entity recognition (NER). The decoder takes the conditioned learned representation produced by the encoder then generates context questions whose answers are spanned in the context.
\par
Experiments on question-answer extraction were conducted by \citep{lewis2021paq}. They introduced Probably Asked Question, a large semi-structured knowledge database of about 65M pairs. The QA database was built using a question generation model and corpora from Wikipedia. To ensure the worthiness of generated pair, they introduced a global-filtering algorithm that ranks the generated pairs. Such a huge QA database could be useful to extend research on question generation and question answering tasks. 

\par
Question Generation by Transformers \citep{kriangchaivech2019question} explores the potential of the vanilla architecture of Transformer to deal with generating questions. Instead of fully encoding the context and answer as they appear in the dataset, some transformations including the change of the named entities by their corresponding NER tags were applied both on the context and answer. 
\par 
A Reinforcement Learning Based Graph-to-Sequence Model was introduced by \citep{chen2020reinforcement} showed a tailored use of reinforcement learning techniques (policy gradient algorithm) for language modeling. They introduced a reinforcement learning-based graph-to-sequence model for question generation, in turn, bring in a deep alignment technique to improve the expressiveness of the encoder.

\section{Model}
Question generation can be identified as sequence transduction consisting in transforming the combination of the context and answer sequences into question. In our setting, the context, the answer and the question are sequences of word respectively identified by $\{x_1^c,...,x_n^c\}$, $\{x_1^a,...,x_m^a\}$ and $\{y_1,...,y_r\}$. Here $n$, $m$, and $r$ appropriately denoted the length of the context, the answer, and the question. Generate answer-aware question consists in generating a question that maximize the probability $P(y_1,...,y_r | X^c, X^a)$
\par 
Previous related work focused on generating questions whose answers are spanned in the context; The extension of this limit was realised by simply modifying the structure of the model’s inputs such that any answer could be attached to a given context rather than only the context spanned sections. An exhaustive test of this extension was not carried out in this work. The model's inputs are formed by the concatenation of context sequence and answer sequence separated by a special token $\{[CLS],x_1^c,...,x_n^c, [SEP],x_1^a,...,x_m,[SEP]\}$. Transformers are universal approximators of sequence to sequence transformations, thanks to the positional encoding and self-attention mechanism, the Transformer encoder will learn the positions of the answer, learn how to establish the relation with the context through the attention mechanism, and finally outputs a better contextual representation of the inputs. The input structure is also affordable to generated open-ended questions.
\par
Drawing inspiration from reinforcement learning-based graph to sequence architecture \citep{chen2020reinforcement},a fully Transformer-based reinforcement learning generator evaluator architecture is introduced to tackle question generation problems. Transfer learning has become an almost de facto technique in NLP like in Computer Vision, thanks to the massive and rich content data available on the internet like C4 (Colossal Clean Crawled Corpus), large models like GPT, BERT, T5 \citep{gpt, devlin2019bert, raffel2020exploring} can be trained through unsupervised or self-supervised learning with a large amount of unlabeled data. More general-purpose pre-training achieves better performance at a fine-tuning step on a downstream task like summarization, text classification, question answering, and language modeling.

\par
T5 was used as the generator. Training language models that merely maximize the likelihood of the output sequence suffers from exposure bias and inconsistency in the training and evaluation performance measurement. This drawback is addressed in this work by using an evaluator that consists of the mixture of the metrics used at the validation step while optimizing the likelihood of the output sequence, the parameters of the models were also optimized according to metrics used at the evaluation time. The evaluator takes as inputs the generated question and the ground truth then outputs a reward score fed into the generator at the training time. To ensure the correctness of the generated question the mixture of BLEU and ELECTRA-discriminator was used as the evaluator, BLEU is used to increase the n-gram overlap between the ground truth and the predicted question, whilst ELECTRA is used for semantic relation. ELECTRA is a Transformer-based model trained with replaced token detection, a self-supervised learning task in which the model learns how to distinguish whether or not a token is corrupted in the input sequence. ELECTRA converges much faster than its predecessor BERT and achieves better performance than BERT on downstream tasks, which points to its ability to produce a better contextual representation of the input sequences. The exploitation of this advantage leads to the building of a semantic evaluator. The reward function used in the adversarial training is defined as follows:
$$r_1(\hat{y}, y) = BLEU(\hat{y}, y)$$
$$r_2(\hat{y}, y) = cos(E_m(\hat{y}), E_m(y)))$$
$$\widetilde{r}(\hat{y},y) = \alpha \times r_1(\hat{y}, y) + (1 - \alpha) \times r_2(\hat{y}, y)$$
$$r(\hat{y}, y) =  \frac{\widetilde{r}(\hat{y},y) + 1 - \alpha}{2 - \alpha}  $$
$\alpha$ is a scalar, $y$ and $\hat{y}$ designed the ground truth and the generated question (using greedy decoding) respectively, $E_m(y)$ is the classification token's ([CLS]) vector representation provided by ELECTRA's encoder, $r(\hat{y}, y)$ refers to the reward score. 
\par
The loss function used to optimize the model is a mixture of the basic likelihood of the output sequence and a modified likelihood which is called here the \textit{reinforcement learning loss}.
The basis component of the loss is merely the log-likelihood of the output sequence :
$$\mathcal{L}_{base}(\hat{y}, y) = - \Sigma_t log P(y_t | y_{<t})$$
The added reinforcement learning loss function is defined as follows : 
$$\mathcal{L}_{rl}(\hat{y}, y) = -(1 - r(\hat{y}, y)) \Sigma_t log P(y_t | y_{<t})$$
The overall loss function is defined as a linear combination of the two loss function components aforementioned:
$$\mathcal{L}(\hat{y}, y) = \gamma \times \mathcal{L}_{base}(\hat{y}, y) + (1 - \gamma) \times \mathcal{L}_{rl}(\hat{y}, y)$$
$\gamma$ is a scalar. 
The reinforcement learning component of the overall loss function behaves as a self-critical part. It tends to lower the objective function when the generated question is semantically close to the ground truth.\citep{rennie2017selfcritical} use of self-critical for image captioning is fully justified, it significantly increased performance of test metrics on MSCOCO task. Generation of insightful questions remains an open research question; the self-critical training paradigm provides one way to address this task. The generation of clarification questions was approached by \citep{rao2019answerbased} with GAN-like architecture where the generator was sequence-to-sequence model and discriminator a utility function to rank the generated question. The higher the rank, the more the answer to the question does not lie in the context provided to the generator.

\begin{figure}[h!]
    \centering
    \includegraphics[scale=0.5]{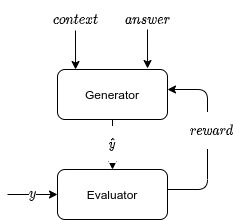}
    \caption{Model architecture}
    \label{fig:model architecture}
\end{figure}

\begin{table*}[t]
    \centering
    \begin{tabular}{llllll}
    \hline
    \textbf{Model} & \textbf{BLEU} & \textbf{Model} & \textbf{BLEU} \\
    \hline
    Transformer \citep{vaswani2017attention} & 3.09  & ASAs2s \citep{Liu_2019} & 16.17\\
    SeqCopyNet \citep{zhou2018sequential}  & 13.02 & G2S$_{sta}$+BERT+RL \citep{chen2020reinforcement} & 18.30 \\
    NQ++ \citep{zhou2017neural} & 13.29 & T5$_{base}$ & 21.31\\
    MPGR+R* \citep{song2018graphtosequence} & 14.71 & T5$_{base}$ + RL$_{base}$(BLEU + Semantic) & 21.52\\
    AFQA \citep{answer-focus-qg} & 15.64 &  T5$_{base}$ + RL$_{base}$(BLEU) & 22.05\\
    s2sa-at-mp-gsa \citep{paragraph-level-qg} & 15.82 & Prophenet \citep{qi2020prophetnet} & \textbf{23.91}\\
    \hline
    \end{tabular}
    \caption{evaluation on test set on BLEU metric}
    \label{tab:evaluation-bleu}
\end{table*}

\begin{table*}[t]
    \centering
    \begin{tabular}{llllll}
    \hline
    \textbf{Model} & \textbf{BLEU} & \textbf{BERTScore} & \textbf{Semantic relation} & \textbf{Logical Agreement} \\
    \hline
    T5$_{base}$ + RL$_{base}$(BLEU + Semantic) & 21.52 & 52.60 & 61.51 & 44.16 \\
    T5$_{base}$ + RL$_{base}$(BLEU) & 22.05 & 52.65 & 61.31 & 42.52 \\
    \hline
    \end{tabular}
    \caption{Semantic evaluation on test set. Semantic relation and Logical Agreement drew from NUBIA}
    \label{tab:evaluation-semantic}
\end{table*}
\begin{table*}[h!]
    \centering
    \begin{tabular}{lllllll}
    \hline
    \textbf{Type} & \textbf{What/Which} & \textbf{Why/How} & \textbf{Where} & \textbf{When} & \textbf{Who} & \textbf{How many}\\
    \hline
    Frequency & 44.16\% & 5.83\% & 4.16\% & 12.49\% & 19.16\% & 14.16\% \\
    Human score & 67.92 & 57.14 & 80 & 100 & 69.56 & 76.47 \\
    Bert score & 50.55 & 48.83 & 46.75 & 64.31 & 56.95 & 53.95 \\
    Semantic relation & 64.80 & 57.95 & 53 & 74.8 & 63 & 57.2 \\
    Logical agreement & 48.63 & 28.91 & 13.07 & 61.88 & 45.75 & 29.65 \\
    \hline
    \end{tabular}
    \caption{Human evaluation and semantic evaluation on T5$_{base}$ + RL$_{base}$(BLEU + Semantic)}
    \label{tab:human-evaluation}
\end{table*}

\section{Experiments}
The training and the inference of the developed architectures were conducted with the benchmark reading comprehension dataset SQUAD v1.1 \citep{rajpurkar2016squad}, which consists of 100,000+ (context, question, answer) triple posed by crowdworkers on a set of Wikipedia articles. The data is originally split into two parts: the training and test. The test set remains unchanged in our experiment and the training set was randomly split into training and development set with the factors 94\% and 6\% respectively.
\par
At the training time, the parameters of the evaluator remained frozen. The hyper-parameter searching was performed on the following hyper-parameters: $\gamma$, $\alpha$, $learning \; rate (lr)$, $batch \; size (bs)$. The ranges used for these parameters were : \\ $0.05<\gamma, \alpha \leq 1$; $10^{-6} \leq lr \leq 10^{-4}$; $bs \in \{16, 32, 64, 128\}$. The scaling of the batch size was carried out via gradient accumulation technique. The best model regarding the loss was obtained with the following hyper-parameters : $bs = 32$, $lr=1.17\times 10^{-5}$, $\alpha = 0.197$, $\gamma = 0.09$.

\section{Results}
Following the previous works, the n-gram-based metric BLEU was used to evaluate the syntactic reconstruction ability of the models. Besides the n-gram based metrics, the semantic side of the generated questions was assessed with the use of semantic-based metrics, namely BERTScore \citep{zhang2020bertscore}, NUBIA \citep{kane2020nubia}. The semantic relation  and logical agreement scores were drawn from NUBIA. The evaluation was performed solely on the test set provided in the original dataset. The table \ref{tab:evaluation-bleu} summarizes the evaluation for the metric BLEU and the table \ref{tab:evaluation-semantic} refers to the evaluation for metric NUBIA.
To ensure that the semantic-based adversarial training effectively improves the performance of the model toward the chosen metric, the following models were independently trained (1) T5$_{base}$ + RL$_{base}$(BLEU), (2) T5$_{base}$ + RL$_{base}$(BLEU + Semantic).

\section{Discussion}
The first fact that could be drawn from the results of table \ref{tab:evaluation-semantic} is the difference of the BLEU score of the two model configurations that have been undertaken namely (i) BLEU-based discriminator and (ii) the blending of BLEU and the semantic component to form the discriminator. Indeed the configuration (i) got a BLEU score greater than the configuration (ii). Tapping into how the discriminators of both structures have been made, this first result conveys that using the model is effectively trained toward improving the performance according to the metric used in the discriminator. That is, the BLEU score obtained with the discriminator only based on BLEU is greater than the score of the discriminator using the mixing of BLEU and the semantic similarity checker.
The second insight is drawn from the analysis of the semantic score of the two configurations. 
Albeit the BLEU score of the configuration (ii) is less than the BLEU score of the configuration (i), the semantic-based scores of the configuration (ii) are commensurable with or even better than the semantic-based scores of the  configuration (i). This reveals a change in the syntactic structure of the generated question while the semantic relatedness between the ground truth remains unchanged or even improved. This second fact corroborates the aforesaid hypothesis stating that the model’s parameters are optimized toward enhancing the performance regarding the metric used in the discriminator component. This observation can be justified by the fact that the values of the hyper-parameters, namely $\alpha, \gamma$  convey that the embedding cosine similarity of the representation outputs by the discriminator has a high contribution to the reward score and the reinforcement learning loss component has a high contribution to the overall loss. 
Both architectures developed outperforms the vanilla architecture of the T5 base, this authenticates the usefulness of the generator-evaluator architecture set to address the question generation. In terms of BLEU  score. The exploration of the large architectures of T5 was not carried out in this work. But instead, the potential of relatively small architecture to perform on par with the larger ones leading the leaderboard.
In addition to automatic evaluation, a binary human-based evaluation of the semantic correlation between the ground truth and the prediction was carried out. 500 questions were randomly sampled from the initial dataset (SQuAD v1.1) and the model used was the T5$_{base}$ + RL$_{base}$(BLEU + Semantic) configuration. The generated questions were classified in different WH-question classes, then for each class, the proportion of questions positively ranked by 5 crowdworkers was calculated. Based on the results in table \ref{tab:human-evaluation}, it is noted that the model hardly generates questions related to explanation whilst it has an affordable performance when the questions are related to time and date. The aim of this human evaluation step is to track the type of questions where the model has a good/bad performance to design a specific discriminator for each type of question in further work.

\bibliographystyle{unsrtnat}
\bibliography{references}  






\end{document}